%% file: main.tex
\documentclass[10pt,twocolumn,letterpaper]{article}

\usepackage[pagenumbers]{cvpr} 

\input{preamble}

\definecolor{cvprblue}{rgb}{0.21,0.49,0.74}
\usepackage[pagebackref,breaklinks,colorlinks,allcolors=cvprblue]{hyperref}

\title{ROAD: Reciprocal-Objective Alignment of Discriminative Semantics for 3D Shape Generation}

\author{
Xiao Luo$^{1}$ \quad
Mingyang Du$^{1}$ \quad
Xin Zhou$^{1}$ \quad
Tianrui Feng$^{1}$\\
Xiwu Chen$^{2}$ \quad
Xiaofan Li$^{3}$ \quad
Jiangning Zhang$^{3}$ \quad
Dingkang Liang$^{1*}$\\[2pt]
$^{1}$Huazhong University of Science and Technology, China\\
$^{2}$Megvii, China
\qquad
$^{3}$Zhejiang University, China\\
{\tt\small
\{dkliang, tianruifeng, xzhou03\}@hust.edu.cn
}\\[2pt]
{\small
\href{https://h-embodvis.github.io/ROAD/}
{\textbf{Project Page}}
}\\[-2pt]
{\small $^{*}$Corresponding author}
}

\begin{document}
\twocolumn[{
  \renewcommand\twocolumn[1][]{#1}

  \maketitle

  \vspace{-0.8em}

  \begin{center}
    \centering
    \includegraphics[width=\textwidth]{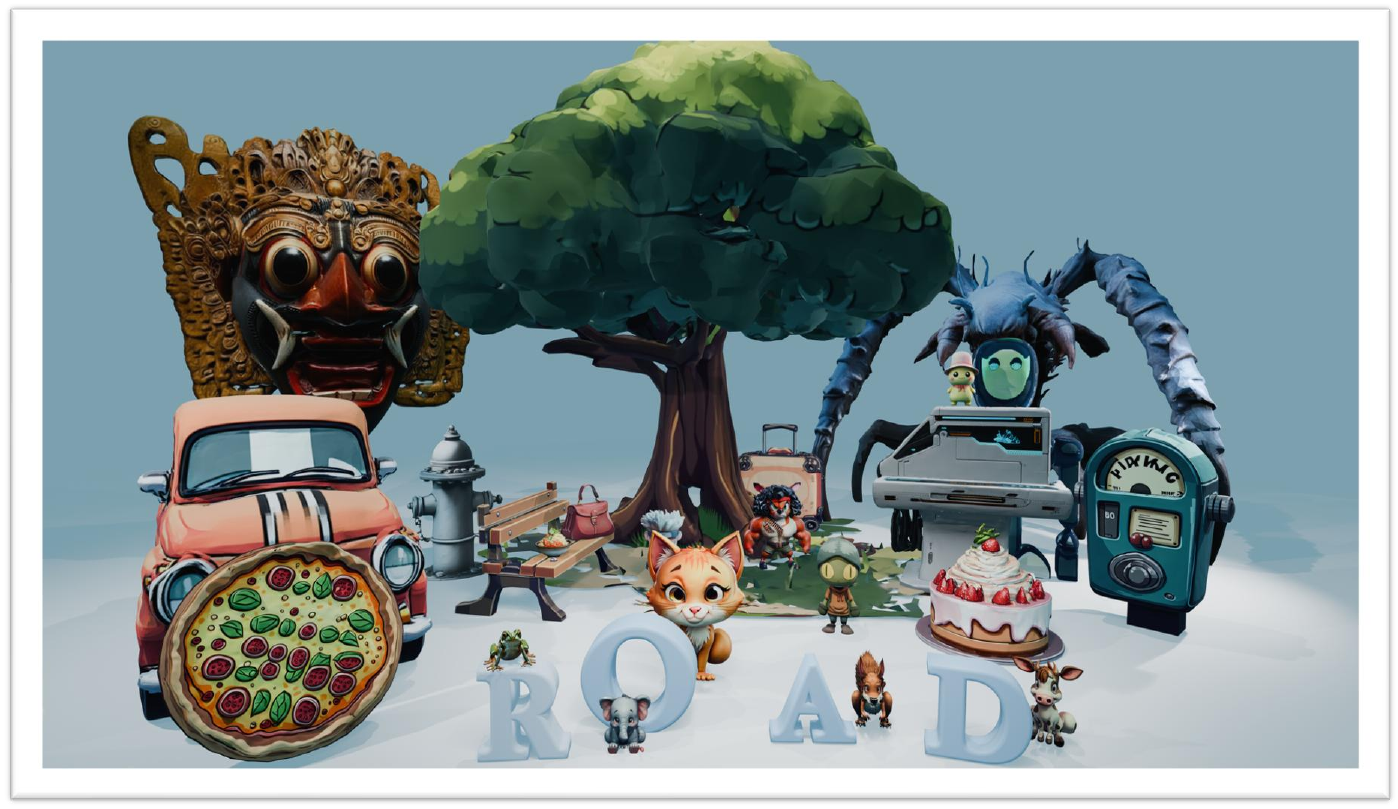}

    \vspace{-0.5em}

    \captionof{figure}{
      High-fidelity 3D assets generated by {\ourmethod}.
      These results showcase the capability of our method to synthesize
      diverse 3D models across various categories and styles with superior
      geometric and textural quality.
    }
    \label{fig:teaser}
  \end{center}

  \vspace{0.5em}
}]

\begin{abstract}
High-fidelity 3D generation predominantly relies on scaling model capacity and data, which incurs prohibitive computational costs. This paradigm typically requires learning geometry from scratch and overlooks the rich semantic and structural priors already encapsulated in discriminative 3D foundation models. We contend that leveraging the profound understanding of the 3D world possessed by these discriminative models can significantly reduce generative cost. To this end, we propose \textbf{\ourmethod}, a framework that reduces the training cost of 3D generation by transferring these rich discriminative priors into diffusion transformers. To address the inherent semantic-structural heterogeneity between generative and discriminative latents, we introduce a reciprocal-objective alignment strategy. This method synergizes Holistic Semantic Condensing to enforce global semantic coherence and Structural Optimal Alignment, which is formulated as a bipartite matching problem to rigorously align microscopic geometric details between disparate latent spaces. The 3D foundation model is only used for training-time supervision of alignment and is not used at inference, incurring no additional inference cost. Compared with the industrial baseline Step1X-3D, the proposed \textbf{\ourmethod} achieves highly competitive generation performance with only 1.5\% of the training data and significantly reduces training costs, effectively reducing the computational overhead of high-fidelity 3D generation. 
Code: \href{https://github.com/H-EmbodVis/ROAD} {https://github.com/H-EmbodVis/ROAD}.
\end{abstract}

\begin{figure*}[t!]
  \centering
  \includegraphics[width=0.96\linewidth]{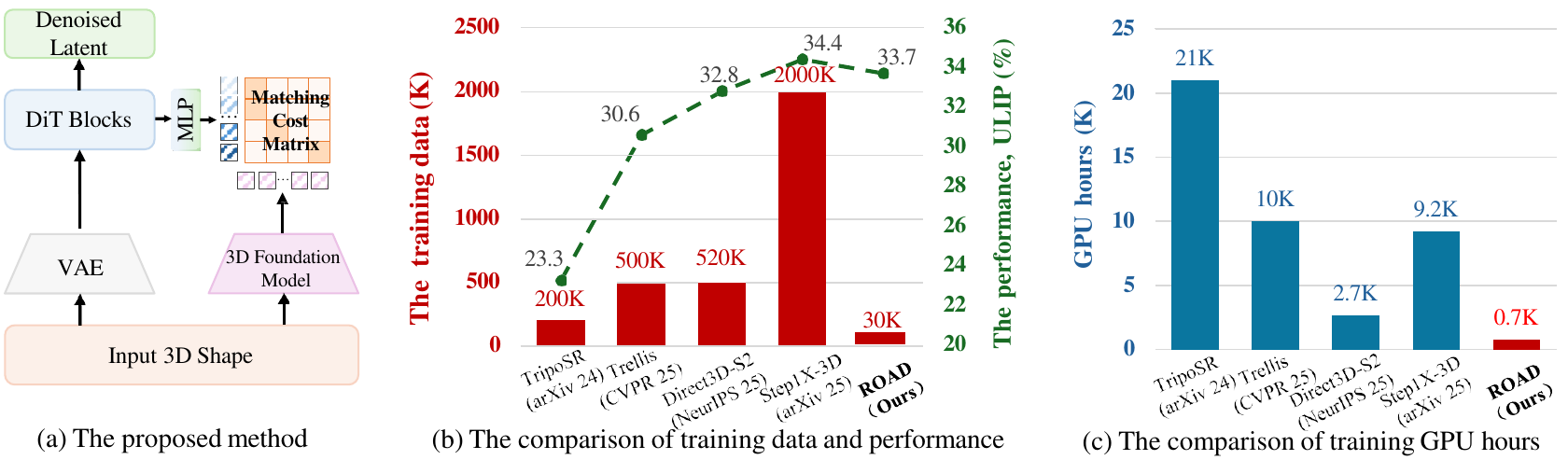}
  \caption{(a) Overview of our proposed {\ourmethod} framework. Instead of relying on resource-intensive, brute-force scaling, we introduce a highly streamlined design tailored for semantic-structural alignment. (b) Computation-performance Pareto frontier. We benchmark our method against prior state-of-the-art works, showcasing a standard-setting trade-off between training data requirements and downstream efficacy. (c) Unprecedented training efficiency. Our method requires the absolute minimum GPU hours among all existing approaches under equivalent or estimated hardware configurations.}
  \label{fig:intro}
  \vspace{-9pt}
\end{figure*}

\section{Introduction}
3D Shape Diffusion Models (3SDMs) have emerged as the dominant paradigm for generating high-fidelity 3D assets. Recent frontiers, such as Hunyuan3D 2.1~\cite{hunyuan3d2025hunyuan3d}, Step1X-3D~\cite{li2025step1x}, and TRELLIS~\cite{xiang2025structured}, have pushed the boundaries of geometric realism. However, their success relies heavily on a scaling strategy, e.g., Step1X-3D consumes 96 $\times$ A100 GPUs over 4 days on 2,000k data. We argue that this paradigm masks a fundamental cognitive inefficiency: these models attempt to learn to generate 3D geometry from scratch, which requires massive resources to establish the mapping between conditions and geometric structures, ignoring the fact that discriminative 3D Foundation Models (3DFMs)~\cite{xue2023ulip,liu2023openshape,zhou2024uni3d,xue2024ulip,liang2025parameter} have encapsulated a profound understanding of the 3D world. Why compel a generative model to relearn the world from raw geometric primitives, when it could inherit the rich, pre-trained structural priors from a discriminative foundation model? 

Recently, using pretrained discriminative foundation models to guide generative training has become an active direction in both image~\cite{leng2025repa} and video~\cite{zhang2025videorepa} generation. By properly integrating the priors of foundation models from their respective modalities, these works have shown remarkable improvements in generative quality and capability. Therefore, we believe that leveraging the large-scale pre-trained knowledge of 3DFMs to bootstrap 3SDMs is a promising direction. Given that these foundation models have acquired universal 3D discriminative representations from billion-scale training, they fundamentally capture the essential features of diverse 3D geometries at the latent level. Injecting these strong priors acts as a cognitive shortcut, potentially allowing shape generative models to achieve high-fidelity synthesis with a fraction of the data.

However, bridging the gap between discriminative priors and generative latents is non-trivial due to the inherent semantic-spatial misalignment. In the domains of image and video generation, a naive or intuitive approach is often adopted to enforce spatial element-wise alignment (i.e., assuming strict positional correspondence between tokens). This strategy proves effective primarily due to the high structural similarity of their encoding frameworks. Specifically, both discriminative and generative models typically process the inputs using identical mechanisms such as shared grid-based patch partitioning, inherently preserving consistent spatial processing relationships and maintaining a strictly aligned order within the token. Yet, our investigation reveals that this strategy is fundamentally flawed in 3D generation due to semantic-structural heterogeneity. 

We attribute this to two factors: first, the permutation-invariant nature of 3D point clouds essentially leads to distinct patch groupings, making deterministic token alignment nearly intractable.  Second, more importantly, 3D foundation models and generative models rely on disparate encoding architectures. Generative models typically aggregate global surface context into sparse latent anchors,  with each token attempting to represent the overall surface information, resulting in tokens that are holistic and position-agnostic. In contrast, 3DFMs tend to encode grouped localized geometric features that maintain strong regional distribution. Consequently, even when processing the identical spatial patch, 3SDMs and 3DFMs often represent entirely disparate information for one encompassing global context and the other capturing local details. This structural-semantic discrepancy renders standard spatial element-wise alignment ineffective, as there is no natural one-to-one correspondence between the two latent sequences.

To unlock the potential of representation alignment in 3D shape generation, we strive to seek a simple yet versatile approach that bridges this inherent mismatch between generative and foundation models. While directly integrating their features to align decoupled high-level semantics might seem like a straightforward solution, it inevitably suffers from a lack of fine-grained constraints, for generative models heavily rely on local details. Given that each token in both models inherently encapsulates geometric information, naturally, we argue that establishing explicit correspondences between these two unordered sets is of paramount importance. The successful application of Hungarian Matching to address similar set-to-set assignment problems in 2D object detection~\cite{carion2020end} serves as a profound inspiration for our approach.

In this paper, we propose \textbf{\ourmethod} (\textbf{R}eciprocal \textbf{O}bjective \textbf{A}lignment of \textbf{D}iscriminative Semantics for 3D Shape Generation), a plug-and-play framework designed to infuse geometric representation of 3DFMs into 3SDMs. Our framework harmonizes the disparate latents and establishes a multi-granularity synergistic alignment, enabling the generative model to simultaneously assimilate macroscopic semantic abstractions and microscopic geometric primitives while preserving its native flexibility.

Specifically, this mechanism efficiently enforces representation coherence through two synergistic pathways: Holistic Semantic Condensing (HSC) and Structural Optimal Alignment (SOA). The HSC employs global anchoring to aggregate disordered tokens into a semantic centroid, compelling the generated latent distribution to align with the macroscopic semantic manifold defined by the foundation model. In contrast, SOA formulates token correspondence as a bipartite matching problem, leveraging the Hungarian algorithm to dynamically match tokens based on semantic similarity for granular alignment. These two reciprocal objectives ensure the effective transfer of features, without disrupting the native generative flow.

We evaluate our approach using the representative open-source framework Step1X-3D~\cite{li2025step1x}. We find that \ourmethod~ substantially reduces the computational cost, as shown in Fig.~\ref{fig:intro}(b)-(c). By acting as a concise semantic bridge, {\ourmethod} achieves performance with significantly fewer parameters and training resources, outperforming several state-of-the-art models. These results demonstrate that semantic alignment is a potent substitute for massive data scaling, offering a path to a more accessible 3D generative community.

Our major contributions are as follows: \textbf{1)} We reveal the inherent representation misalignment between 3D foundation models and generative models, identifying this structural discrepancy as the primary obstacle to effective prior transfer. \textbf{2)} We propose \ourmethod, a framework that bridges these disparate latent spaces via a stratified alignment mechanism across both semantic abstraction and structural detail, ensuring rigorous prior injection. \textbf{3)} We demonstrate that effective prior transfer significantly reduces data and training costs while maintaining competitive fidelity, thereby realizing the efficiency of high-quality 3D generation.

\section{Related Work}
\subsection{3D Shape Generation}
\label{subsec:3D shape generation}

Recently, 3D shape generation, a core task in computer vision, has emerged as a prominent research direction, as it demonstrates the capacity to liberate 3D modeling from labor-intensive manual techniques through automated generation. However, various technical limitations persist. To lift lower-dimensional conditions to global 3D features, early optimization-based frameworks~\cite{poole2022dreamfusion,wang2023score,wang2023prolificdreamer, shi2023mvdream, yu2024homugan, yang2023single} construct latent 3D representations from multi-view images, which are constrained by slow convergence and high rendering costs. Later works transitioned to efficient feed-forward networks trained on diverse representations, including point clouds~\cite{vahdat2022lion,jun2023shap,nichol2022point,li2021hsgan,zhou2025recurrent}, meshes~\cite{siddiqui2024meshgpt,chan2022efficient,wan2024cad}, and hybrid features~\cite{shue20233d,tochilkin2024triposr,hu2024efficientdreamer}. Fine-grained point-cloud generation and enhancement have also been explored through hierarchical graph modeling and frequency-aware generative learning~\cite{li2021hsgan,liu2022pufagan}, further demonstrating the importance of preserving local geometric details in 3D synthesis. However, these models do not alleviate the reliance on extensive multi-view images, still imposing a heavy training burden.

Modern approaches typically employ Variational Autoencoders (VAEs) to compress geometry into compact latent spaces. This strategy successfully aggregates high-dimensional features, allowing for the derivation of full 3D shape structures from single-view inputs. While state-of-the-art models like Tripo series~\cite{tochilkin2024triposr,li2025triposg}, Hunyuan3D~\cite{hunyuan3d2025hunyuan3d,lai2025hunyuan3d}, Step1X-3D~\cite{li2025step1x}, and TRELLIS~\cite{xiang2025structured} achieve remarkable fidelity, they nevertheless primarily rely on expanding model capacity and data, which incurs substantial computational demands and resource overhead, making them increasingly inaccessible to the broader academic community.

Unlike these approaches that rely on brute-force scaling, we investigate empowering generative frameworks by transferring rich priors from discriminative 3D Foundation Models (3DFMs), achieving high-quality generation with significantly reduced computational overhead.

\subsection{3D Foundation Models}
\label{subsec:3D foundation models}
3D foundation models extract geometric and semantic information from unordered point clouds. Early approaches use shared MLPs for independent feature extraction~\cite{qi2017pointnet,rao2020global,qian2022pointnext}, whereas later advancements leverage local structural contexts, inter-region relations, high-order geometric dependencies, and global representations~\cite{cheng2021net,zhang2020hypergraph,lin2023meta,sun2024x,wang2024gpsformer}. These pioneering models establish a robust paradigm for processing unstructured 3D data, effectively transforming raw point clouds into high-dimensional latent representations enriched with deep semantic cues.

Significant strides have been made in 3D representation learning, driven by diverse architectures including the Point Transformer series~\cite{zhao2021point,wu2022point,wu2024point} and PointMamba~\cite{liang2024pointmamba}, as well as self-supervised frameworks like PointMAE~\cite{pang2023masked}, Recon~\cite{qi2023contrast}, and masked point-cloud representation learning~\cite{wang2024rethinking}. Moreover, cross-modal models such as PointCLIP~\cite{zhang2022pointclip} and ShapeLLM~\cite{qi2024shapellm} have successfully scaled discriminative capabilities by aligning 3D data with rich semantic spaces. These models leverage massive datasets to encapsulate robust structural and semantic abstractions. Instead of limiting 3DFMs to discrimination, we propose transferring their rich semantic and structural priors (e.g., Uni3D~\cite{zhou2024uni3d}) to generative frameworks. 

Despite their potential, the direct integration of strong discriminative features from 3D models poses significant hurdles~\cite{he2026tango3d}. {\ourmethod} bridges this representational mismatch, enabling the seamless injection of latent features into generative models. By aligning generative latents with these robust features, we bootstrap the generation process, significantly enhancing fidelity.

\subsection{Representation Alignment}

Recently, visual representation alignment (e.g., REPA~\cite{yurepresentation}, U-REPA~\cite{tian2025u}, REPA-E~\cite{leng2025repa}) has successfully enhanced the semantic information of image generative models by injecting representation information of foundation models, token by token, significantly boosting the training efficiency and convergence speed of generative models. Subsequent work SRA~\cite{jiang2025no} replaces representation alignment with self-alignment. IREPA~\cite{singh2025matters}, and REG~\cite{wu2025representation}, enrich semantic guidance to refine REPA, while VideoREPA~\cite{zhang2025videorepa} and GeometryForcing~\cite{wu2025geometry} extend these methods to video generation. However, these approaches require a strict spatial, element-wise, one-to-one correspondence across the encoded tokens. In 3D shape generation, the semantic-structural heterogeneity between 3SDMs and 3DFMs disrupts the correlation, which presents an alignment challenge. 

Consequently, in this paper, we propose a novel framework, {\ourmethod}, to explore representation alignment within generative models for 3D generation, thereby lowering the entry barriers to high-quality 3D content creation.

\section{Preliminary}
\subsection{3D Foundation Model Paradigm}
To handle unordered point clouds, an input point cloud 
$\mathbf{Q}=\{\mathbf{q}_i\}_{i=1}^{N_p}$ with optional point attributes 
$\mathbf{A}=\{\mathbf{a}_i\}_{i=1}^{N_p}$ is first partitioned into $G$ localized patches. For the $g$-th patch centered at $\mathbf{c}_g \in \mathbb{R}^{3}$, its local geometric encoding is formulated as:
\begin{equation}
f_g =
\operatorname{Pool}_{\mathbf{q}_i \in \mathcal{N}(\mathbf{c}_g)}
\left(
\operatorname{MLP}
\left(
[\mathbf{q}_i-\mathbf{c}_g;\mathbf{a}_i]
\right)
\right),
\end{equation}
where $\mathcal{N}(\mathbf{c}_g)$ denotes the neighboring points around $\mathbf{c}_g$, and $\mathbf{a}_i$ represents the attribute of point $i$, such as normals or colors. The operator $[\cdot;\cdot]$ denotes feature concatenation.

The patch features are then projected and organized into a transformer input sequence:
\begin{equation}
\mathbf{F}^{0}
=
\left[
\mathbf{f}_{\mathrm{cls}}+\mathbf{p}_{\mathrm{cls}};
\left\{
\operatorname{Proj}(f_g)+\operatorname{PosEnc}(\mathbf{c}_g)
\right\}_{g=1}^{G}
\right],
\end{equation}
where $\mathbf{f}_{\mathrm{cls}}$ and $\mathbf{p}_{\mathrm{cls}}$ denote the learnable classification token and its positional embedding, respectively. $\operatorname{Proj}(\cdot)$ is a linear projection layer, and $\operatorname{PosEnc}(\cdot)$ denotes the positional encoding derived from patch centers. After the transformer encoder, we denote the extracted patch-level discriminative features as
\begin{equation}
\mathbf{Y}=\{\mathbf{y}_i\}_{i=1}^{M}\in \mathbb{R}^{M\times D'},
\end{equation}
which serve as the 3D foundation-model priors used in our framework. Since point-cloud patches are produced by sampling and grouping operations, $\mathbf{Y}$ is treated as an unordered set of local semantic-structural tokens rather than a sequence with canonical ordering.

\subsection{Hungarian Matching Paradigm}
Given two unordered sets 
$\mathcal{U}=\{\mathbf{u}_i\}_{i=1}^{K}$ and 
$\mathcal{V}=\{\mathbf{v}_j\}_{j=1}^{K}$, 
Hungarian Matching provides a standard solution for establishing one-to-one correspondences without relying on predefined element order. For each pair $(\mathbf{u}_i,\mathbf{v}_j)$, a matching cost $\mathcal{D}_{ij}$ is computed according to the task requirement. The general formulation can be written as:
\begin{equation}
\mathcal{D}_{ij}
=
\lambda_{\mathrm{cls}}
\mathcal{L}_{\mathrm{cls}}(\mathbf{u}_i,\mathbf{v}_j)
+
\lambda_{\mathrm{reg}}
\mathcal{L}_{\mathrm{reg}}(\mathbf{u}_i,\mathbf{v}_j),
\end{equation}
where $\mathcal{L}_{\mathrm{cls}}$ and $\mathcal{L}_{\mathrm{reg}}$ denote the classification and regression costs, and $\lambda_{\mathrm{cls}}$ and $\lambda_{\mathrm{reg}}$ are balancing coefficients.

The optimal assignment is obtained by searching for a permutation $\sigma^{*}$ that minimizes the total matching cost:
\begin{equation}
\sigma^{*}
=
\operatorname*{argmin}_{\sigma \in \mathfrak{S}_{K}}
\sum_{i=1}^{K}
\mathcal{D}_{i,\sigma(i)},
\end{equation}
where $\mathfrak{S}_{K}$ denotes the set of all permutations over $K$ elements and $\sigma(i)$ denotes the mapped value for $i$-th input. After obtaining $\sigma^{*}$, the matched loss is computed as:
\begin{equation}
\mathcal{L}_{\mathrm{match}}
=
\frac{1}{K}
\sum_{i=1}^{K}
\mathcal{L}
\left(
\mathbf{u}_i,
\mathbf{v}_{\sigma^{*}(i)}
\right).
\end{equation}
In this way, Hungarian Matching removes the ambiguity caused by unordered elements and enables permutation-invariant set-to-set supervision. In our method, we instantiate this paradigm with feature-similarity costs to align generative tokens with 3D foundation-model tokens.

\begin{figure*}[ht]
  \centering
  \includegraphics[width=0.95\linewidth]{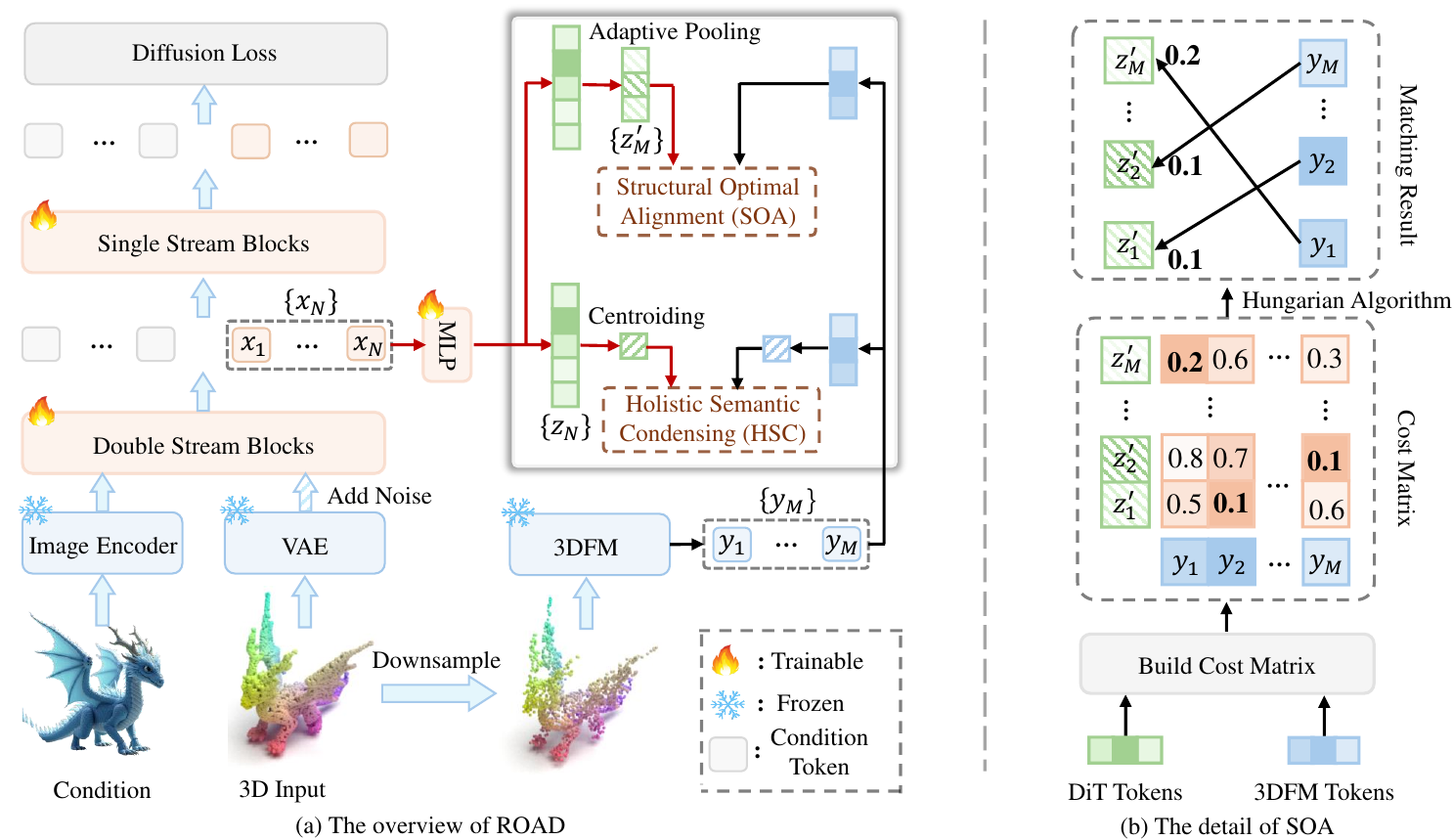}
  \caption{Overview of the proposed {\ourmethod}. (a) During training, the frozen 3D Foundation Model (3DFM) provides semantic priors to guide the diffusion process through complementary holistic consistency and similarity-based semantic alignment. (b) Details of Structural Optimal Alignment (SOA). We build a cost matrix between DiT tokens and 3DFM tokens, and then apply the Hungarian algorithm to obtain token-level matching results for semantic alignment. }
  \label{fig:pipeline}
  \vspace{-9pt}
\end{figure*}

\section{Our Method}
\label{sec:method}

While representation alignment significantly bolsters the efficacy of generative models, a straightforward adaptation of successful 2D methodologies fails to adequately address the specialized requirements of 3D synthesis. In this section, we present \textbf{\ourmethod}, a framework that scales down the cost of high-fidelity generation by injecting the rich priors of discriminative 3D Foundation Models (3DFMs) into 3D Shape Diffusion Models (3SDMs). As illustrated in Fig.~\ref{fig:pipeline}, our framework operates by integrating frozen 3DFMs  (e.g., Uni3D) as a semantic supervisor alongside the standard generative framework. Here, the frozen 3DFM is employed only during training to compute the proposed alignment constraints. To inject discriminative priors without compromising the generative model's flexibility, we introduce a reciprocal-objective alignment stream comprising two synergistic components: Holistic Semantic Condensing (HSC) for global semantic coherence, and Structural Optimal Alignment (SOA) for fine-grained geometric fidelity. By employing this dual-alignment strategy, the potent discriminative priors from the foundation model are effectively integrated into the generative process. Consequently, this approach significantly reduces training overhead while enabling the synthesis of high-quality 3D assets.

\subsection{Overall Framework}
\label{subsec:overall_framework}

To evaluate the feasibility of our approach, we adopt the state-of-the-art Step1X-3D~\cite{li2025step1x} framework as our baseline. This framework generally comprises a SDF Variational Autoencoder (VAE) for geometric compression and a Diffusion Transformer (DiT) for latent generation.

\textbf{SDF Variational Autoencoder.} To compress continuous geometry into a compact representation, the framework utilizes an SDF-VAE~\cite{chen2025dora}. Given a point set $\mathcal{P}$, the encoder $E_\phi$ employs a cross-attention to aggregate global context onto $N$ sparse spatial anchors sampled (key points) via Farthest Point Sampling (FPS), yielding latent tokens $\mathbf{X} = \{ \mathbf{x}_i \}_{i=1}^N \in \mathbb{R}^{N \times D}$, where $D$ denotes the feature dimension. A decoder $D_\theta$ then queries these tokens to predict the Signed Distance Field (SDF) values for surface reconstruction. Such an SDF-VAE compresses complex geometric spaces into dense SDF fields, where each token processes global features rather than localized aggregation.

\textbf{3D Diffusion Transformer.} The generative backbone is a Multi-modal Diffusion Transformer (MM-DiT)~\cite{esser2024scaling} designed to model the latent distribution, which is trained under a flow-matching paradigm. Operating directly on $\mathbf{X}$, the model learns a velocity field over continuous latent trajectories conditioned on semantic context $\mathbf{c}$. Specifically, given a clean latent $\mathbf{x}_0$ encoded by the VAE and a Gaussian noise latent $\mathbf{x}_1 \sim \mathcal{N}(\mathbf{0}, \mathbf{I})$, we construct an interpolated latent
\begin{equation}
    \mathbf{x}_t = (1-t)\mathbf{x}_0 + t\mathbf{x}_1, \quad t \in (0,1),
\end{equation}
and train the network to predict the corresponding flow velocity. The architecture incorporates double-stream and single-stream blocks to facilitate interaction between geometric tokens and condition features, optimized via the diffusion objective:
\begin{equation}
    \mathcal{L}_{\text{diff}} = 
    \mathbb{E}_{\mathbf{x}_0, \mathbf{x}_1, t, \mathbf{c}}
    \left[
    \left\|
    \left(\mathbf{x}_1 - \mathbf{x}_0\right)
    -
    G_\theta(\mathbf{x}_t, t, \mathbf{c})
    \right\|^2
    \right].
\end{equation}
In the overall training pipeline, the VAE first encodes the 3D shape into latent tokens, which are then linearly interpolated with Gaussian noise and fed into the DiT alongside conditional inputs to learn the latent flow. Our ROAD framework intervenes in this flow by injecting discriminative priors from a frozen 3D foundation model into DiT blocks via the proposed reciprocal-objective constraints.

\subsection{Investigation: Semantic-Structural Heterogeneity}
\label{subsec:misalignment}

Before elaborating on the dual-alignment methodology, we first identify the fundamental barrier hindering the direct transfer of priors: the semantic-structural heterogeneity between generative and discriminative representations. To fully characterize this misalignment, it is essential to analyze the inherent disorder of point clouds alongside the structural discrepancies between these two representational paradigms.

For densely sampled point clouds, existing methods predominantly rely on random sampling or Farthest Point Sampling (FPS) to select cluster centers, which are then organized into token sequences. However, this approach inherently introduces stochasticity into the sequence composition. Consequently, in tasks requiring precise token-to-token alignment, 3D representations lack a standardized serialization protocol, unlike the well-defined grid-based tokenization prevalent in image preprocessing. A straightforward approach would be to forcibly fix the point sequences of the generative model with those of the foundation model. However, this strategy encounters a subsequent challenge, as the two paradigms employ fundamentally different encoding strategies for point representation.

\begin{figure}[t]
  \centering
  \includegraphics[width=0.96\linewidth]{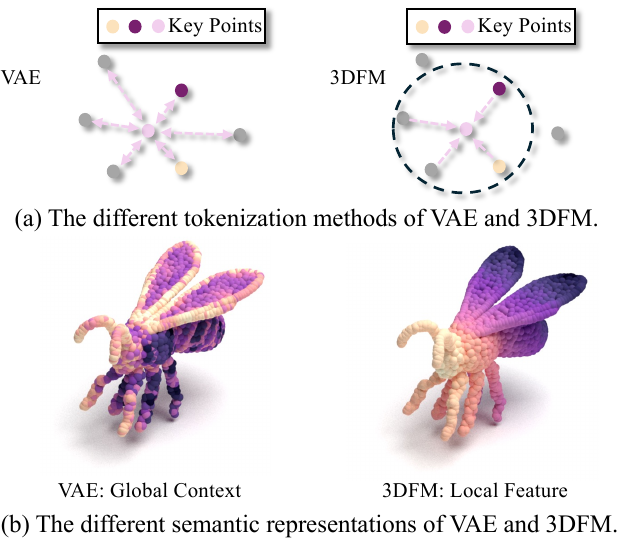}
  \caption{Analysis of semantic misalignment. (a) A showcase of different tokenization methods. (b) Feature projections confirm the representation gap.}
  \label{fig:representation_gap}
    \vspace{-9pt}
\end{figure}

Initially, we observe that the feature sequence encoded by the SDF-VAE exhibits permutation invariance. Specifically, the decoder can reconstruct the correct geometry even if the latent tokens are shuffled. This implies that the VAE hidden states do not strictly encode sequential correspondence. We attribute this to the tokenization mechanism shown in Fig.~\ref{fig:representation_gap}(a), where the VAE employs cross-attention to aggregate context from the entire point set $\mathcal{P}$ onto sparse anchors, whereas standard 3DFMs typically utilize explicit grouping operations (e.g., KNN) to encode geometry from local features. Consequently, DiT tokens behave as order-agnostic latent descriptors of the surface rather than strictly localized patch descriptors, yet 3DFMs' tokens maintain strong spatial correlation. This discrepancy is indicated by the feature visualization in Fig.~\ref{fig:representation_gap}(b), where points with similar color encode homogeneous information. The visualization confirms that the generative sequence lacks a fixed spatial correspondence with the discriminative sequence, rendering rigid element-wise alignment fruitless and necessitating our proposed dual-alignment strategy.

\subsection{Reciprocal-Objective Prior Alignment Mechanism}
\label{subsec:dual_alignment}

To resolve the semantic-structural heterogeneity identified above, we propose a unified mechanism that harmonizes representations at both the global and local levels. This process involves feature projection, holistic condensing, and structural bipartite matching.

\textbf{Feature Projection and Normalization.} Informed by recent research in representation alignment, we begin by performing feature projection. Let $\mathbf{Y} = \{ \mathbf{y}_i \}_{i=1}^M \in \mathbb{R}^{M \times D'}$ denote the features extracted from the frozen 3DFM, where $M$ represents the number of latent tokens and $D'$ denotes the feature dimensionality. We extract the intermediate features $\mathbf{X}$ from 3SDMs (see Sec.~\ref{subsec:layer_selection}) and employ a lightweight MLP projector $\Phi_{\text{MLP}}$ to map $\mathbf{X}$ into the 3DFM's latent space, yielding $\mathbf{Z} = \{ \mathbf{z}_i \}_{i=1}^N = \Phi_{\text{MLP}}(\mathbf{X})\in \mathbb{R}^{N \times D'}$.
To prevent numerical discrepancies from biasing the alignment, we apply $L_2$ normalization to yield $\hat{\mathbf{z}}_i$ and $\hat{\mathbf{y}}_j$. These preparations significantly enhance the stability of the alignment and encourage the model to focus on more informative features in subsequent alignment.

\textbf{Holistic Semantic Condensing (HSC).}
Given the semantic-structural heterogeneity between the two models, direct element-wise alignment is unstable. Therefore, we first attempt to enforce alignment at the macroscopic level. We utilize Holistic Semantic Condensing to aggregate disordered tokens into unified semantic prototypes via global average pooling:
\begin{equation}
    \mathbf{c}_z = \frac{1}{N} \sum_{i=1}^{N} \hat{\mathbf{z}}_i, \quad \mathbf{c}_y = \frac{1}{M} \sum_{j=1}^{M} \hat{\mathbf{y}}_j.
\end{equation}
This holistic anchoring method resolves the unordered nature of 3D representations, which treats tokens as geometry integrals, overlooking the discrepancies between them. The HSC loss is then defined as the cosine distance between these centers, compelling the generative model to orient towards the foundation model's semantic centroids:
\begin{equation}
    \mathcal{L}_{\text{HSC}} = 1 - \text{cos}(\mathbf{c}_z, \mathbf{c}_y).
\end{equation}

\begin{figure}[t]
  \centering
  \includegraphics[width=0.85\linewidth]{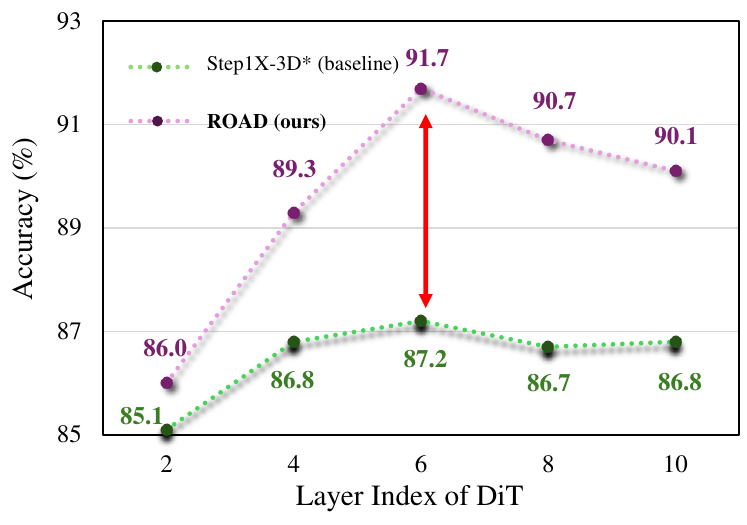}
  \caption{Classification result on ModelNet40. Our model significantly enhances the semantic information of the representations.}
  \label{fig:classification}
    \vspace{-9pt}
\end{figure}

\begin{table*}[htbp]
\centering
\small
\setlength{\tabcolsep}{1.5mm}
\caption{Efficiency benchmarking against state-of-the-art 3D generative models. We compare {\ourmethod} with leading methods trained on both proprietary and public datasets. {\ourmethod} achieves highly competitive performance while utilizing substantially fewer training data (only 30k) and significantly fewer parameters, effectively facilitating high-fidelity 3D generation.}
\vspace{-9pt}
\label{tab:main_results}
\begin{tabular}{lcccccc}
\toprule
Model & Venue & Params. & Public Data & Private Data & Uni3D-Score $\uparrow$ & ULIP-Score $\uparrow$ \\
\midrule
\multicolumn{7}{c}{\textit{\textbf{Commercial Models}}} \\
\midrule
Hunyuan3D-2.1~\cite{hunyuan3d2025hunyuan3d} & arXiv 25 & 3.3B & 100k & \checkmark & 33.3 & 33.1 \\ 
Step1X-3D~\cite{li2025step1x} & arXiv 25 & 1.3B & 800k & 1200k & 33.5 & 34.4 \\
\midrule
\multicolumn{7}{c}{\textit{\textbf{Academic Models}}} \\
\midrule
TripoSR~\cite{tochilkin2024triposr} & arXiv 24 & 0.4B & 200k & \xmark & 24.6 & 23.3 \\
TRELLIS~\cite{xiang2025structured} & CVPR 25 & 2.0B & 500k & \xmark & 26.8 & 30.6 \\
Hi3DGen~\cite{ye2025hi3dgen} & ICCV 25 & 4.2B & 170k & 700k & 31.0 & 32.8 \\
Direct3D-S2~\cite{wu2025direct3d} & NeurIPS 25 & 2.0B & 520k & \xmark & 31.8 & 32.8 \\
\ourmethod  ~(\textbf{ours}) & - & 0.65B & 30k & \xmark & \textbf{32.3} & \textbf{33.7} \\
\bottomrule
\end{tabular}
\vspace{-9pt}
\end{table*}

\textbf{Structural Optimal Alignment (SOA).}
While HSC ensures global semantic coherence, the condensing process inevitably discards spatial details. Prior point-cloud generation and enhancement studies have shown that fine-grained geometric cues are crucial for preserving structural fidelity~\cite{li2021hsgan,liu2022pufagan}. To improve fine-grained structural information without rigid element-wise alignment, we formulate the local mapping as a bipartite matching problem. Specifically, we apply small-scale adaptive average pooling to $\mathbf{Z}$, obtaining $\mathbf{Z}' = \{ \mathbf{z'}_i \}_{i=1}^M \in \mathbb{R}^{M \times D'}$. This step establishes the equal-cardinality condition required for one-to-one Hungarian matching, while also serving as a lightweight denoising operation to suppress training-induced perturbations in the latent tokens. Then, we compute a pairwise cost matrix $\mathbf{C} \in \mathbb{R}^{M \times M}$ based on the cosine distance between all token pairs:
\begin{equation}
    \mathbf{C}_{i,j} = 1 - \text{cos}(\mathbf{z}'_i, \mathbf{y}_j).
\end{equation}
We seek an optimal permutation $\pi^*$ from the set of all possible permutations $\{\pi\}$ that minimizes the total matching cost using the Hungarian algorithm:
\begin{equation}
    \pi^* = \operatorname*{argmin}_{\pi} \sum_{i=1}^{M} \mathbf{C}_{i, \pi(i)}.
\end{equation}
where $\pi(i)$ indicates the mapped index of the $i$-th element under the permutation $\pi$.

Consequently, following the mapping provided by $\pi^*$, we align the $i$-th token from 3SDM with the $\pi^*(i)$-th token of 3DFM using the same cosine distance as in $\mathbf{C}$. The SOA loss is then formulated as the average cosine distance between matching tokens:
\begin{equation}
    \mathcal{L}_{\text{SOA}} = \frac{1}{M} \sum_{i=1}^{M} \mathbf{C}_{i, \pi^*(i)}.
\end{equation}
With the implementation of SOA, the model is empowered to capture microscopic information missed in HSC while overcoming the limitations of semantic discrepancies in rigid element-wise alignment, thereby enforcing fine-grained local constraints and providing it with more concrete semantic details.

\textbf{Training Objective.}
The final training objective combines the standard diffusion loss with our dual constraints:
\begin{equation}
    \mathcal{L} = \mathcal{L}_{\text{diff}} + \lambda_1 \mathcal{L}_{\text{HSC}} + \lambda_2 \mathcal{L}_{\text{SOA}},
\end{equation}
where $\lambda_1 = \lambda_2 = 0.5$ are weighting coefficients used to balance the contribution of each loss term.

\subsection{Analysis on Semantic Layer Selection}
\label{subsec:layer_selection}

Recent research indicates that varying alignment depths exert distinct influences on model performance. To identify the optimal layer for feature alignment that yields richer semantic priors, we investigate the generative backbone's internal representation hierarchy. Specifically, we utilize the Step1X-3D*\footnote{\textit{* indicates that the Step1X-3D DiT is trained from scratch in our configuration (see Sec.~\ref{details})}} as our probe subject, extracting intermediate features from the frozen DiT blocks and training lightweight linear classifiers on the ModelNet40 dataset. The results, visualized in Fig.~\ref{fig:classification}, reveal that the $6^{th}$ layer reaches peak performance. Following this, accuracy gradually declines, suggesting a transition from semantic representation to abstract geometric encoding. Consequently, we explicitly target this layer for our alignment, ensuring that our discriminative priors are injected into the most receptive layers.

\section{Experiment}

\subsection{Implementation details}\label{details}

We adopt the state-of-the-art 3D diffusion model, Step1X-3D~\cite{li2025step1x}, as our generative baseline. Note that to improve efficiency, we streamline the model architecture by reducing the parameter count by $\sim$50\%, termed Step1X-3D*. Specifically, while the original Step1X-3D comprises 12 double-stream blocks and 24 single-stream blocks, our pruned variant Step1X-3D* utilizes 6 and 12, respectively. We employ the pre-trained Uni3D-G~\cite{zhou2024uni3d} as our discriminative foundation model for alignment. During training, we only update the DiT parameters and the projection MLP. This DiT is trained entirely from scratch without loading any pre-trained weights. Based on semantic analysis (see Sec.~\ref{subsec:layer_selection}), we extract the intermediate features from the $6^{th}$ layer (double-stream) to perform the proposed alignment. All experiments are conducted on 8 NVIDIA A100 GPUs, with a per-GPU batch size of 16 and a learning rate of $1 \times 10^{-3}$. The model is trained for 130k iterations, taking approximately 3.5 days to converge.

\begin{table}[t]
\centering
\small
\setlength{\tabcolsep}{0.1mm}
\caption{The comparisons between our method and other visual representation alignment methods.}
\vspace{-9pt}
\label{tab:ablation_models}
\begin{tabular}{lccc}
\toprule
Alignment methods & Venue& Uni3D-Score $\uparrow$ & ULIP-Score $\uparrow$ \\
\midrule
Baseline & - & 31.3 & 32.5\\
REPA~\cite{yurepresentation} &ICLR 25 & 30.8 & 31.3\\
REG~\cite{wu2025representation} & NeurIPS 25& 30.1 & 28.2 \\
SRA~\cite{jiang2025no} & ICLR 26& 29.4 & 30.1 \\
\ourmethod ~(\textbf{ours}) &- &  \textbf{32.3} & \textbf{33.7}\\
\bottomrule
\end{tabular}
\vspace{-9pt}
\end{table}

\subsection{Datasets and Evaluation Metrics}
\label{datasets}

We construct our training set by only randomly sampling 30,000 assets from the publicly available Objaverse~\cite{deitke2023objaverse} dataset. The data processing pipeline strictly adheres to the protocols established by Step1X-3D. In a departure from the original setup, we render only 12 views for each 3D asset, as opposed to the 20 views utilized in the baseline. For evaluation, we curate a robust test set comprising 160 unseen assets, strictly ensuring asset-level non-overlap with the training data. Following previous methods~\cite{li2025step1x,hunyuan3d2025hunyuan3d,wu2025direct3d}, we assess the semantic consistency using standard CLIP-Score metrics, reporting results derived from Uni3D-G~\cite{zhou2024uni3d} and ULIP~\cite{xue2023ulip}.

\subsection{Main results}

\textbf{Comparison with state-of-the-art methods.} We perform a comprehensive evaluation against a wide range of frontier 3D generative models, classifying them into commercial solutions and academic frameworks. As detailed in Table~\ref{tab:main_results} and Fig.~\ref{fig:intro}, {\ourmethod} demonstrates exceptional performance-efficiency trade-offs. Specifically, when compared to TripoSR~\cite{tochilkin2024triposr}, {\ourmethod} delivers substantial improvements in geometric quality and semantic consistency, outperforming it by significant margins in both Uni3D-Score (+7.7) and ULIP-Score (+10.4). Even against large-scale latent models such as TRELLIS~\cite{xiang2025structured} (2.0B params, 500k data), {\ourmethod} (0.65B, 30k) yields superior generative quality. Notably, regarding data efficiency, {\ourmethod} surpasses the recent Direct3D-S2~\cite{wu2025direct3d} (ULIP-Score: 32.8) while utilizing only 5.7\% of its training data (30k vs. 520k) and one-third of its parameters, validating the economy of representation alignment in 3D shape generation.

Furthermore, {\ourmethod} challenges models reliant on massive proprietary datasets, including commercial giants like Step1X-3D~\cite{li2025step1x} and Hunyuan3D~\cite{hunyuan3d2025hunyuan3d}, as well as the recent academic model Hi3DGen~\cite{ye2025hi3dgen}. While these methods require million-scale private data (e.g., 700k--1200k) and prohibitive compute (e.g., 96 $\times$ GPUs), {\ourmethod} achieves parity with a mere public dataset and can be trained on a single node with just 8 A100 GPUs. By slashing resource requirements while maintaining exquisite quality, {\ourmethod} downscales high-fidelity 3D generation for the academic community.

\textbf{Comparison with other representation alignment methods.} To better elucidate the challenge of directly transferring 2D representation-alignment strategies to the 3D domain, we compare {\ourmethod} with recent visual representation alignment methods~\cite{yurepresentation,wu2025representation,jiang2025no}. These methods are originally designed for image/video generation, where tokens usually follow regular grid structures or relatively stable positional correspondences. To adapt them to 3D generation, we further enforce exact coordinate correspondence between spatial point tokens to reduce the confounding randomness introduced by Farthest Point Sampling (FPS). As illustrated in Table~\ref{tab:ablation_models}, these transferred alignment strategies lead to noticeable performance degradation, whereas {\ourmethod} achieves consistent improvements. For rigid one-to-one alignment strategies such as REPA and REG, the enforced coordinate correspondence can disrupt the holistic surface representations of generative latents, resulting in inferior generation quality. For self-alignment strategies such as SRA, we conjecture that the relatively small generative model produces noisy intermediate representations during early training, making self-alignment less reliable before the model converges. In contrast, {\ourmethod} introduces a reciprocal-objective alignment paradigm that bridges the semantic-structural heterogeneity of 3D features through both holistic semantic guidance and similarity-based structural matching, effectively overcoming the limitations of rigid positional alignment.

\begin{figure}[t!]
  \centering
  \includegraphics[width=0.86\linewidth]{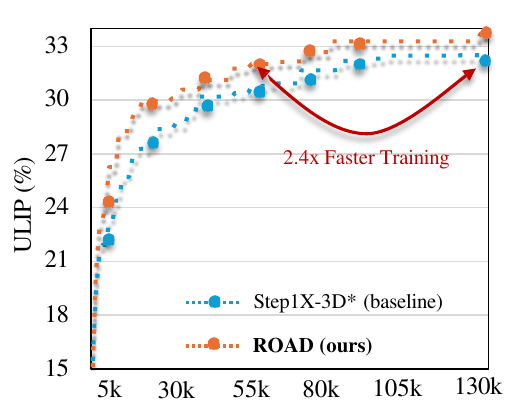} 
  \vspace{-9pt}
  \caption{Convergence Analysis. Our method achieves the baseline's peak performance 2.4$\times$ faster (at 55k vs. 130k iterations).}
  \vspace{-9pt}
  \label{fig:convergence}
\end{figure}

\textbf{Training convergence analysis.}
As illustrated in Fig.~\ref{fig:convergence}, the proposed {\ourmethod} demonstrates significantly faster convergence compared to the Step1X-3D* baseline. Quantitatively, our model reaches the baseline's peak performance (ULIP-Score $\approx$ 32.5\%) at approximately 55k iterations, whereas the baseline requires over 130k iterations, effectively yielding a 2.4$\times$ acceleration. This speedup confirms that injecting discriminative priors effectively constrains the optimization search space, preventing the model from slowly learning semantic structures.

\subsection{Qualitative Analysis}

\begin{figure*}[t]
  \centering
  \includegraphics[width=0.91\linewidth]{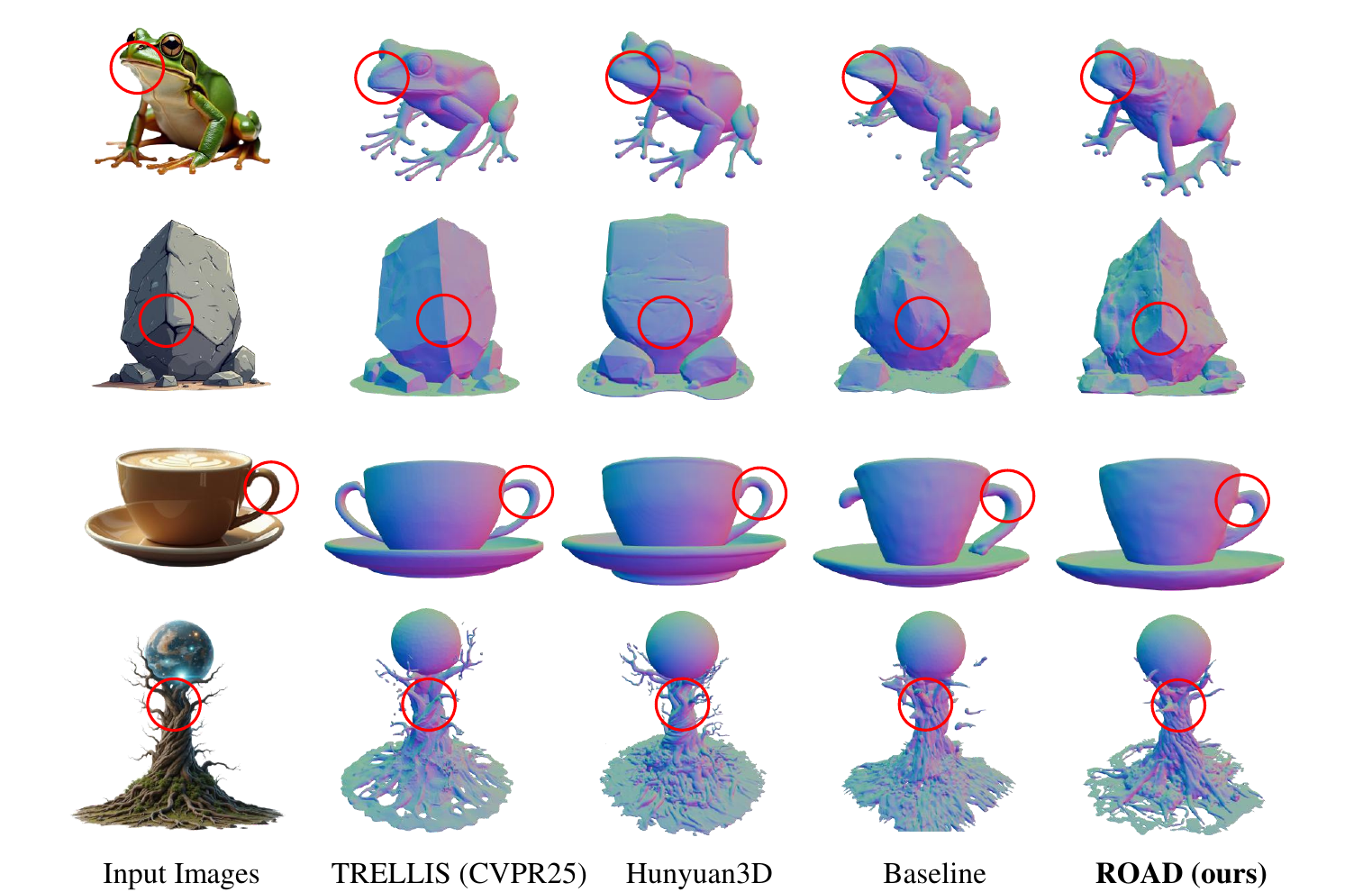}
  \vspace{-9pt}
  \caption{{\ourmethod} is comparable to TRELLIS (an academic benchmark) in geometric fidelity and achieves visual quality comparable to Hunyuan3D (an industrial paradigm), while effectively mitigating the structural artifacts observed in the baseline.}
  \label{fig:visual_compare_geomertry}
  \vspace{-9pt}
\end{figure*}

\begin{figure*}[t]
  \centering
  \includegraphics[width=0.91\linewidth]{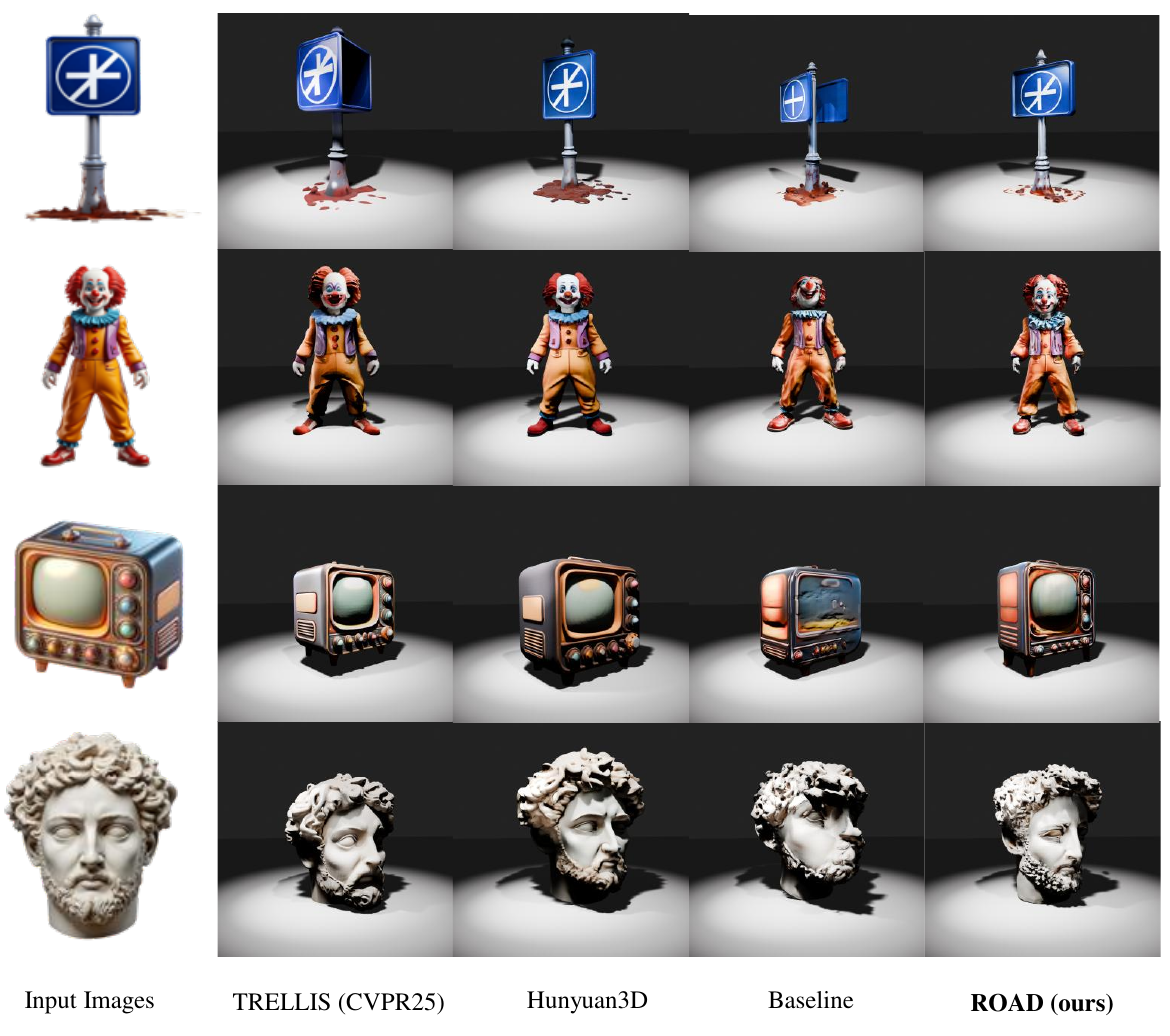}
  \vspace{-9pt}
  \caption{Our {\ourmethod} surpasses TRELLIS (academic models) in detail fidelity and achieves visual parity with Hunyuan3D (industrial models), while effectively rectifying the structural artifacts observed in the baseline.}
  \label{fig:visual_compare}
  \vspace{-9pt}
\end{figure*}

\textbf{Generation fidelity compared with other models.} As illustrated in Fig.~\ref{fig:visual_compare_geomertry} and Fig.~\ref{fig:visual_compare}, ROAD demonstrates improved generation quality compared to academic baselines (e.g., TRELLIS), delivering sharper geometric details and more faithful structures while mitigating over-smoothing artifacts. Most notably, {\ourmethod} achieves visual parity with industrial-scale models (e.g., Hunyuan3D), producing high-fidelity assets that are comparable in terms of structural integrity and realism, despite utilizing orders of magnitude less data. Furthermore, ROAD significantly rectifies the structural distortions and semantic inconsistencies inherent in the unaligned baseline, validating the feasibility of injecting discriminative priors and efficacy of our alignment strategy. 

\begin{figure}[t]
  \centering
  \includegraphics[width=0.96\linewidth]{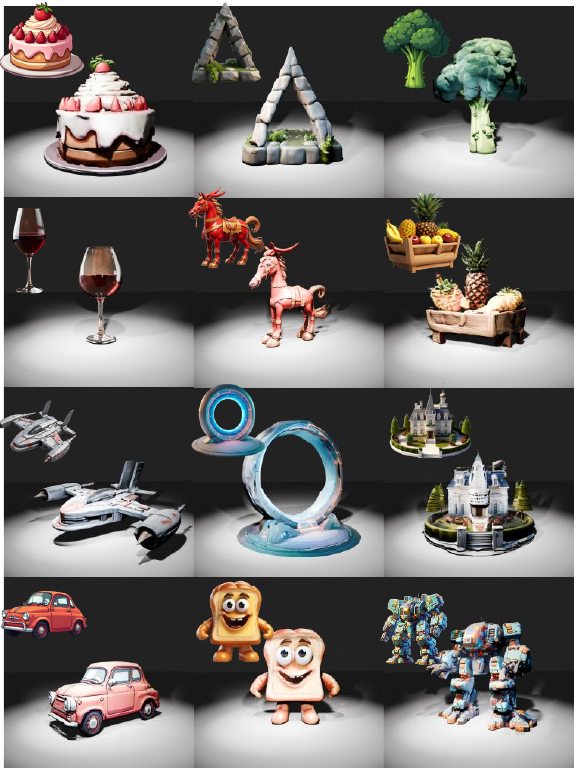}
  \vspace{-9pt}
    \caption{Visualizations across diverse asset categories demonstrate the model's robust generative capacity under various stylized conditions.}
  \label{fig:demo}
  \vspace{-9pt}
\end{figure}

\textbf{Visualizations of diverse 3D assets.} To showcase the capability of our approach in generating high-fidelity 3D assets, we condition the generation process on a set of diverse input images collected from multiple sources. The final synthesized results are presented in Fig~\ref{fig:demo}. Impressively, our model delivers exceptional generative performance across a wide range of object categories, thereby validating the robust generalization of {\ourmethod}.

\subsection{Model Transferability}
To verify the generalizability of our method, we also conduct experiments across diverse architectures, including \textbf{CraftsMan}\cite{li2025craftsman3d} and \textbf{TRELLIS}\cite{xiang2025structured}. For simplifying the experimental pipeline, unless specified otherwise, both models are trained and evaluated exclusively on the compact dataset declared in Sec.~\ref{datasets}.

\textbf{Transfer to Alternative SDF-Based Generators.} Although CraftsMan shares a similar SDF latent representation with Step1X-3D, the native CraftsMan architecture eschews the sharp surface embedding employed by the latter. Furthermore, during the diffusion stage, CraftsMan relies on a conventional DiT backbone rather than the joint single-stream and dual-stream design found in Step1X-3D. Utilizing CraftsMan as a testbed, we investigate whether our proposed method maintains its generalizability under a framework constrained by less descriptive representations. 

Specifically, we implement a straightforward alignment on the 6th DiT block of CraftsMan. Prior literature posits that early layers in generative models exhibit a stronger affinity for high-level semantic abstractions. Consequently, enforcing representational alignment at an early stage yields a more pronounced guidance effect. The quantitative results are detailed in Table~\ref{tab:combined_results}. Our proposed method yields compelling results when integrated into the CraftsMan framework, successfully validating its applicability to fundamental generative baselines. 

\textbf{Generalization to Structured Sparse-Latent Architectures.} TRELLIS adopts a novel structured latent representation to encode 3D geometric shapes. Specifically, this structured formulation voxelizes the underlying geometry, treating each occupied voxel block as an individual token. These tokens are subsequently organized via a specialized serialization protocol, ensuring that each token inherently encapsulates localized geometric information coupled with a distinct spatial part. However, this sparse voxel-based tokenization strategy still exhibits a pronounced discrepancy from the point-based tokenization paradigms utilized in foundation models. To validate the generalizability and robustness of our method with structured spatial latents, we apply our ROAD framework to TRELLIS. Concurrently, to streamline our training pipeline, we deliberately deploy our method within Stage 1, which exerts more profound impacts on geometric quality.

We strictly follow the method established in the original TRELLIS work to process our 30k dataset. As demonstrated by the quantitative evaluations compiled in Table~\ref{tab:combined_results}, our proposed method is successfully validated on the TRELLIS framework. Specifically, the variant enhanced by our method consistently outperforms the baseline trained solely on the identical dataset without alignment. Furthermore, our model achieves competitive performance that is comparable to the original TRELLIS model trained under standard conditions. This consistent improvement demonstrates that ROAD can also benefit sparse structured latents and further underscores the cross-architecture stability of our framework.

\begin{table}[t]
\centering
\small
\caption{Results of CraftsMan and TRELLIS.}
\vspace{-9pt}
\setlength{\tabcolsep}{10pt}
\label{tab:combined_results}
\begin{tabular}{ccc}
\toprule
Variant  & Uni3D-Score $\uparrow$ & ULIP-Score $\uparrow$ \\
\midrule
\multicolumn{3}{l}{\textit{Craftsman Model}} \\
\midrule
W/o ROAD & 26.7 & 30.5 \\
W/ ROAD  & 27.9 & 31.7 \\
\midrule 
\multicolumn{3}{l}{\textit{TRELLIS Model}} \\
\midrule
W/o ROAD & 26.6 & 29.8 \\
W/ ROAD  & 27.0 & 30.6 \\
\bottomrule
\end{tabular}
\vspace{-9pt}
\end{table}

\subsection{Ablation Studies}
Unless otherwise stated, the following experiments are conducted on the Step1X-3D* baseline to validate the effectiveness of our method.

\textbf{The effect of reciprocal-objective alignment.} We first verify the necessity of our dual-stream design. As presented in Table~\ref{tab:key_components}, deploying Holistic Semantic Condensing (HSC) or Structural Optimal Alignment (SOA) individually yields marginal improvements. This suggests that aligning solely at the macroscopic or microscopic level is insufficient to capture the foundation priors fully. However, their integration triggers a synergistic effect, boosting the ULIP-Score significantly to 33.7 (+1.2). We attribute this to their complementary nature: HSC acts as a global stabilizer, ensuring the generation aligns with the correct semantic manifold (what to generate), while SOA refines the local topology via bipartite matching (how to generate details). {\ourmethod} effectively harmonizes these two granularities. 

\begin{table}[t]
\centering
\setlength{\tabcolsep}{4.mm}
\small
\caption{Impact of alignment components. We evaluate the contribution of Holistic Semantic Condensing (HSC) and Structural Optimal Alignment (SOA).}
\vspace{-9pt}
\label{tab:key_components}
\begin{tabular}{cccc}
\toprule
HSC & SOA & Uni3D-Score $\uparrow$ & ULIP-Score $\uparrow$\\
\midrule
- & - & 31.3 & 32.5 \\
\checkmark & - & 31.4 & 32.8 \\
- & \checkmark & 31.7 & 33.2 \\
\checkmark & \checkmark & \textbf{32.3} & \textbf{33.7} \\
\bottomrule
\end{tabular}
\vspace{-9pt}
\end{table}

\begin{table}[t]
\centering
\setlength{\tabcolsep}{0.6mm}
\small
\caption{Effectiveness of different matching metrics for SOA. The experiments are equipped with HSC.}
\vspace{-9pt}
\label{tab:ablation_costs}
\begin{tabular}{lccc}
\toprule
Basis & Matching Metric & Uni3D-Score $\uparrow$ & ULIP-Score $\uparrow$ \\
\midrule
Spatial & Point Coordinate $L_2$ & 30.9 & 32.4 \\
\midrule
\multirow{2}{*}{Semantic} & Latent Feature $L_2$  & 31.6 & 33.0\\ 
 & Latent Feature Cosine & \textbf{32.3} & \textbf{33.7} \\
\bottomrule
\end{tabular}
\vspace{-9pt}
\end{table}

\begin{table}[t]
\centering
\small
\setlength{\tabcolsep}{2mm}
\caption{Impact of 3D Foundation Model (3DFM) Capacity. }
\vspace{-9pt}
\label{tab:3DFM}
\begin{tabular}{lccc}
\toprule
Variant & Params & Uni3D-Score $\uparrow$ & ULIP-Score $\uparrow$ \\
\midrule
W/o 3DFMs  & - & 31.3 & 32.5 \\
W/ 3DFMs & 0.3B & 31.9 & 32.7 \\
W/ 3DFMs & 1B & \textbf{32.3} & \textbf{33.7} \\
\bottomrule
\end{tabular}
\vspace{-9pt}
\end{table}

\textbf{Analysis on matching strategy: why semantics matter?} A critical design choice in SOA is the metric used for bipartite matching. Table~\ref{tab:ablation_costs} reveals that employing spatial matching based on keypoint coordinates leads to performance degradation. This empirical failure corroborates our hypothesis in Sec.~\ref{subsec:misalignment} that tokens lack fixed spatial correspondence due to the permutation-invariant nature of the VAE. Consequently, forcing alignment based on physical coordinates disrupts the generative model's native latent topology. In contrast, Semantic Matching successfully bridges this gap by identifying correspondence based on feature content. Specifically, using Latent Feature Cosine similarity yields the optimal performance, verifying that {\ourmethod} aligns geometric details by respecting the semantic-structural heterogeneity between the two models.

\textbf{Impact of foundation model capacity.} We investigate the influence of the pre-trained 3D Foundation Model's (3DFM) scale on prior transfer effectiveness, as shown in Table~\ref{tab:3DFM}. It can be observed that as the model size decreases, the performance gains introduced by alignment gradually diminish. This suggests that weaker discriminators may provide insufficient guidance, resulting in less stable intermediate representations in the generative model during training. Conversely, scaling up to larger foundation models leads to progressive improvements, confirming that a sufficiently strong discriminator is essential for capturing the rich semantic priors needed to guide the generative process.

\begin{table}[t]
\centering
\small
\setlength{\tabcolsep}{0.1mm}
\caption{Exploring the Feasibility of Aligning Diverse Modalities.}
\vspace{-9pt}
\label{tab:modalities}
\begin{tabular}{lccc}
\toprule
Alignment modalities & Venue & Uni3D-Score $\uparrow$ & ULIP-Score $\uparrow$ \\
\midrule
Baseline & - & 31.3 & 32.5\\
DINOv2~\cite{oquab2023dinov2} & TMLR 23 & 29.6 & 30.2\\
VGGT~\cite{wang2025vggt} & CVPR 25& 30.4 & 30.0 \\
Uni3D~(\textbf{ours})  &- &  \textbf{32.3} & \textbf{33.7}\\
\bottomrule
\end{tabular}
\vspace{-9pt}
\end{table}

\textbf{Which modality of representational information is effective?} To identify the optimal representational information for 3D Shape Diffusion Models (3SDMs), we explore foundation models from various modalities, including DINOv2, VGGT, and Uni3D (i.e., 2D, 2.5D, and 3D). As shown in Table~\ref{tab:modalities}, aligning with DINOv2 and VGGT leads to suboptimal performance. By contrast, our approach effectively aligns with native 3D Foundation Model (3DFM), obtaining significantly better results. This indicates that 3SDMs inherently rely on concrete geometry information. Consequently, {\ourmethod} elegantly injects the priors from the 3DFM, generating high-fidelity 3D assets. Specifically, 2D visual features mainly capture appearance-level semantics and lack a comprehensive understanding of complete 3D geometry, making them insufficient for guiding structured shape synthesis. While 2.5D representations provide stronger geometric cues, they are still derived from view-dependent observations and may not fully characterize the continuity and global topology of object surfaces. Consequently, forcibly injecting such mismatched priors can interfere with the generative capacity of 3SDMs. In contrast, native 3D foundation models directly encode spatial structures and surface-level geometric semantics, providing more compatible and informative priors for enhancing 3D generative modeling.

\textbf{Effect of sampling density.} Since sampling density directly affects the representational completeness and noise level of 3D point clouds, we evaluate our method with different numbers of sampled points as inputs. The quantitative results are presented in Table~\ref{tab:density}. We observe that a moderate sampling density achieves the best performance in capturing the representational information of the foundation model. We attribute this to the balance between geometric completeness and noise accumulation. Overly sparse point clouds may fail to fully characterize the underlying 3D geometry, whereas excessively dense point clouds can introduce redundant or noisy local details that disturb feature alignment. Therefore, selecting an appropriate point density strikes an optimal balance, preserving fine-grained details while avoiding redundant noise.

\begin{table}[t]
\centering
\small
\setlength{\tabcolsep}{2mm}
\caption{Effect of Sampling Density. }
\vspace{-9pt}
\label{tab:density}
\begin{tabular}{cccc}
\toprule
Num of Points & Density & Uni3D-Score $\uparrow$ & ULIP-Score $\uparrow$ \\
\midrule
-  & - & 31.3 & 32.5 \\
4096 & Small & 31.9 & 33.4 \\
10000 & Middle & \textbf{32.3} & \textbf{33.7} \\ 
20000 & Large & 31.4 & 32.4 \\
\bottomrule
\end{tabular}
\vspace{-9pt}
\end{table}

\begin{table}[!t]
\centering
\small
\setlength{\tabcolsep}{1.5mm}
\caption{Ablation study on different alignment layers.}
\label{tab:layers}
\begin{tabular}{lcc}
\toprule
Target Layer & Uni3D-Score $\uparrow$ & ULIP-Score $\uparrow$ \\
\midrule
\multicolumn{3}{l}{\textit{Shallow-to-Mid Range (Layers 1--6)}} \\
\midrule
Layer 2 (Double-Stream)& 31.4 & 32.8 \\
Layer 4 (Double-Stream) & 31.3 & 32.5 \\
Layer 6 (Double-Stream) & \textbf{32.3} & \textbf{33.7} \\
\midrule
\multicolumn{3}{l}{\textit{Mid-to-Deep Range (Layers 6--18)}} \\
\midrule
Layer 8 (Single-Stream)& 32.1 & 33.2 \\
Layer 10 (Single-Stream)& 31.4 & 32.7 \\
\bottomrule
\end{tabular}
\end{table}

\textbf{Discussion on the diversity of alignment layers.} The Multi-modal Diffusion Transformer is composed of Double-Stream Blocks (DSB) and Single-Stream Blocks (SSB). Notably, the DSB handles the geometric and conditional image elements independently, whereas the SSB concatenates these two modalities into a unified joint feature. To investigate which feature alignment strategy yields superior performance, we conduct alignments across different block types as well as at various depths within the same block architecture. As illustrated in Table~\ref{tab:layers}, the 6th layer of the Double-Stream blocks yields superior performance. We argue that feature alignment places greater emphasis on semantic intensity. Thus, aligning at layers with stronger representations significantly influences the overall performance, with inter-modal differences having a negligible impact.

\section{Conclusion}
This paper presents \textbf{\ourmethod}, a framework designed to make 3D shape generation efficient by injecting robust priors from discriminative foundation models. We bridge the semantic gap via a reciprocal-objective strategy comprising Holistic Semantic Condensing (HSC) and Structural Optimal Alignment (SOA). This approach effectively harmonizes the unordered nature of 3D latents, ensuring precise alignment without compromising generative flexibility. Comprehensive evaluations demonstrate that {\ourmethod} achieves industrial-tier fidelity with exceptional data efficiency. We hope this work establishes a new paradigm for efficient and accessible 3D generative research.

\textbf{Limitation.} The generation quality is inherently bounded by the semantic capacity of the foundation model. Future work could explore integrating more advanced instructors to further push the boundaries of generation fidelity.

{
    \small
    \bibliographystyle{ieeenat_fullname}
    \bibliography{main}
}


\end{document}

%% file: preamble.tex
\usepackage{pifont}
\usepackage{multirow}
\usepackage{graphicx}
\usepackage{capt-of}

\newcommand{\ourmethod}{ROAD}
\newcommand{\xmark}{\ding{55}}%

